\documentclass[]{spie}  

\usepackage{amsmath,amsfonts,amssymb}
\usepackage[colorlinks=true, allcolors=blue]{hyperref}

\usepackage{algorithmic}
\usepackage{algorithm}
\usepackage{array}
\usepackage{textcomp}
\usepackage{stfloats}
\usepackage{url}
\usepackage{verbatim}
\usepackage{graphicx}
\usepackage{cite}
\usepackage{color}
\usepackage{booktabs}
\usepackage{xcolor}
\usepackage{multirow}
\usepackage{enumitem}
\definecolor{citecolor}{HTML}{0071bc} 
\usepackage{float}
\usepackage{xcolor}
\definecolor{SeaGreen4}{RGB}{0,205,102} 
\definecolor{SlateBlue}{RGB}{106,90,205} 
\definecolor{DarkRed}{RGB}{178,34,34} 
\usepackage{times}
\usepackage{soul}
\usepackage{url}
\usepackage[switch]{lineno}
\usepackage{amssymb}
\usepackage{pifont}
\newcommand{\cmark}{\ding{51}}%
\newcommand{\xmark}{\ding{55}}%
\usepackage{makecell}
\usepackage{colortbl}
\definecolor{mygray}{gray}{.9}
\definecolor{mypink}{rgb}{.99,.91,.95}
\definecolor{mycyan}{cmyk}{.3,0,0,0}
\usepackage{booktabs}

\title{Point-Voxel Absorbing Graph Representation Learning for Event Stream based Recognition}

\author[a]{Yuxiang Zhang}
\author[a]{Chengguo Yuan}
\author[a]{Xiao Wang}
\author[a]{Bo Jiang} 
\author[a]{Jin Tang}

\affil[a]{School of Computer Science and Technology, Anhui University, Hefei 230601, China.}

\authorinfo{Further author information: (Send correspondence to Xiao Wang)\\ 
A.A.A.: E-mail: xiaowang@ahu.edu.cn}

\begin{document} 
\maketitle

\begin{abstract}
Sampled point and voxel methods are usually employed to downsample the dense events into sparse ones. After that, one popular way is to leverage a graph model that treats the sparse points/voxels as nodes and adopts graph neural networks (GNNs) to learn the representation of event data. Although good performance can be obtained, however, their results are still limited mainly due to two issues. (1) Existing event GNNs generally adopt the additional max (or mean) pooling layer to summarize all node embeddings into a single graph-level representation for the whole event data representation. However, this approach fails to capture the importance of graph nodes and also fails to be fully aware of the node representations. (2) Existing methods generally employ either a sparse point or voxel graph representation model which thus lacks consideration of the complementary between these two types of representation models. To address these issues, in this paper, we propose a novel dual point-voxel absorbing graph representation learning for event stream data representation. To be specific, given the input event stream, we first transform it into the sparse event cloud and voxel grids and build \emph{dual absorbing graph models} for them, respectively. Then, we design a novel absorbing graph convolutional network (AGCN)  for our dual absorbing graph representation and learning. The key aspect of the proposed AGCN is its ability to effectively capture the importance of nodes and thus be fully aware of node representations in summarizing all node representations through the introduced absorbing nodes. Finally, the event representations of dual learning branches are concatenated together to extract the complementary information of two cues. The output is then fed into a linear layer for event data classification. Extensive experiments on multiple event-based classification benchmark datasets fully validated the effectiveness of our framework. Both the source code and pre-trained models will be released at: \url{https://github.com/Event-AHU/AGCN_Event_Classification}. 
\end{abstract}


\section{Introduction}  \label{sec:intro}
 
\begin{figure} 
\center
\includegraphics[width=6.5in]{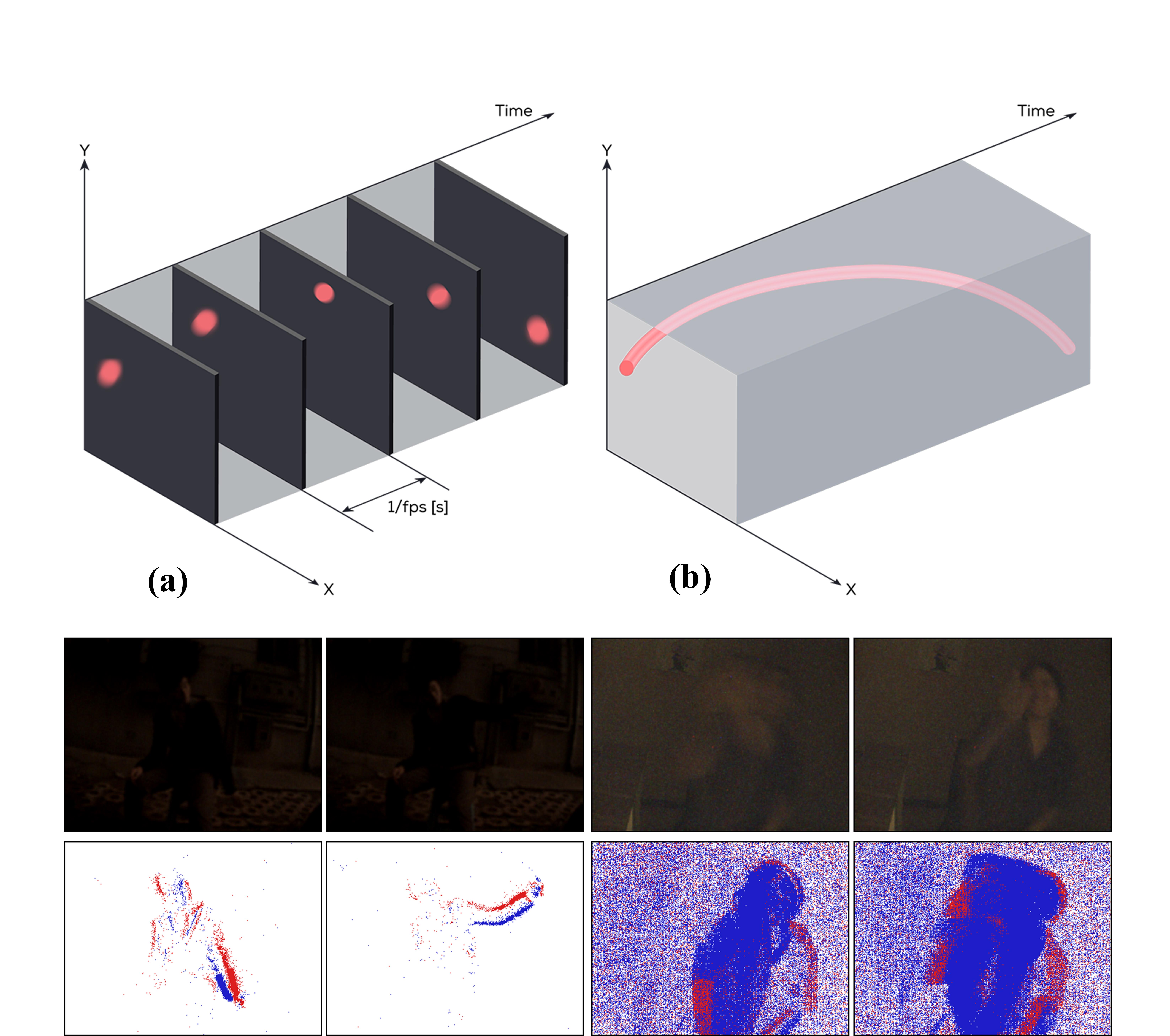}
\caption{The comparison between the imaging principle of (a) frame-based camera and (b) Event camera. Note that, the frame based camera captures the target object in a synchronous video frame, while the Event camera detects the motions in a pulse that is delivered asynchronously. It shows significant advantages in the fast motion and high dynamic range, as shown in the following two bottom rows. Event stream are stacked into images for better visualization.  
}  
\label{firstIMG}
\end{figure}

Event-based vision has drawn more and more attention in recent years. 
It has been widely exploited in both high-level (such as object detection~\cite{mueggler2017fast, li2022retinomorphic}, visual tracking~\cite{wang2021visevent, zhu2022learningGraphTracking, ramesh2020etld}) and low-level (video frame interpolation~\cite{tulyakov2021timelens, chen2022residual}, scene reconstruction~\cite{zhu2021neuspike}, image enhancement~\cite{jiang2023eventLIEnhance}, stereo vision~\cite{uddin2022USeventStereo}) computer vision tasks.  Different from the RGB camera which records the scene into video frames in a synchronous way, each pixel in the event camera is triggered asynchronously by saving an event point if and only if the variation of intensity exceeds the given threshold. 
The event camera shows several advantages or features, such as {high dynamic range}, {low energy-consumption}, and {dense temporal resolution but sparse spatial resolution} \cite{gallegoevent}. Therefore, it performs well even in low-illumination, overexposure, and fast-motion scenarios. A comparison example between the imaging principle and resulting images of the RGB camera and event camera is shown in Fig.~\ref{firstIMG}. One can find that the event stream captures the spatial contour and temporal motion information well compared with  RGB camera.

In this paper, we  focus on the event stream-based classification. 
After reviewing existing works, 
we have the following observations. To be specific, the early researchers stack the asynchronous event stream into synchronous image-like representations~\cite{wang2019evGait} to make full use of off-the-shelf deep models for image/video processing. However, such simple transformations may lose the dense temporal information of the event stream. Recently, current event-based recognition models usually treat the event stream $(x, y, t)$ as point cloud $(x, y, z)$~\cite{wang2019spaceCloud, schaefer2022aegnn, wang2021eventgait3Dgraph}, or divide the event into non-overlapping voxel grids~\cite{xie2022vmv, deng2021evvgcnn, xie2023eventVoxelformer, deng2023dynamicGNN}. Then, they build a graph by treating each point or voxel grid as the node and adopt a Graph Neural Network (GNN) to learn the structured information on the graph. For example, Li et al. propose a graph-based event classification framework that processes the data event-by-event, termed SlideGCN~\cite{LiYijin_2021_ICCV}. Simon et al. propose an event-processing paradigm, termed AEGNN~\cite{schaefer2022aegnn}, which can handle events as evolving spatio-temporal graphs. VMV-GCN~\cite{xie2022vmv} is proposed by Xie et al. which presents a voxel-wise graph learning model for event-based classification.

Although these models work well on simple event-based recognition datasets, however, we believe their performances may be still limited due to the following three issues: 
\textbf{First}, existing event GNNs generally adopt an additional max (or mean) pooling layer to summarize all node embeddings into a single, graph-level representation for event data representation, which fails to capture the importance of graph nodes and also fails to be fully aware of the node representations. 
\textbf{Second}, 
previous works usually adopt 
regular GCNs~\cite{kipf2016semi, bi2020graph,li2021graph} for event graph learning whose receptive field is usually limited due to deeper GCNs usually lead to the over-smoothing issue. Thus, they usually fail to capture the long-range dependencies of different nodes for event data representation. 
\textbf{Third}, 
existing methods usually down-sample the dense event stream into the monotonic event representations (e.g., sparse point, voxel grid), to achieve efficient processing. 
Obviously, these event representations are complementary to each other. For example, sparse point representation mainly focuses on capturing the
distribution of event units 
while voxel can model the 
spatio-temporal relationships more effectively. 
However, to our knowledge, 
existing methods generally either employ a center point or voxel representation model which thus lacks consideration of the complementary between them.

To address the aforementioned issues, in this paper, we propose a novel effective dual absorbing graph representation framework for event-based recognition. 
To be specific, given the input dense event stream, we first down-sample it into the sparse event stream and meanwhile divide it into non-overlapping voxel grids~\footnote{We keep the valid ones according to the number of event points via top-k selection.} respectively.
Then, we build  dual absorbing graph models for
the point and voxel stream separately, each of which includes all sparse point/voxel nodes as well as a particular absorbing node. 
After that, we propose to design a novel absorbing graph convolutional network (AGCN) for our absorbing
graph representation and learning. 
The main benefits of the proposed
AGCN model are three aspects. 
(i) It can effectively capture the importance of
event nodes in learning the graph-level representation via the introduced absorbing node. 
(ii) The absorbing node in AGCN can adaptively absorb (or aggregate) the information of all event nodes, thus allowing for summarizing all node representations more effectively than traditional pooling layer~\cite{simonovsky2017dynamic}. 
(iii) In AGCN, each node aggregates the message from both its neighbors and the absorbing node. 
Since the absorbing node  provides 
a message
bottleneck to encode the global information, AGCN can preserve the local and global structures simultaneously for better learning of graph representation. 
Finally, 
we concatenate the outputs of dual AGCN branches together to extract the complementary information
of two streams and fed it into a linear layer for event recognition. The overview of our learning framework is illustrated in Fig.~\ref{framework}.

To sum up, the main contributions of this paper are as follows: 

\noindent $\bullet$ We propose a novel absorbing graph model for event data representation.  It can  capture the importance of nodes and thus be fully aware of all node representations when summarizing them for graph-level representation. It also conveys global information in each node's message aggregation and thus can capture the global information more effectively in each node's representation. 

\noindent $\bullet$ We design a novel absorbing graph convolutional network (AGCN) to capture both local and global structures for better learning of the proposed absorbing graph representation. 

\noindent $\bullet$ We propose a  dual-stream graph representation learning framework for event-based recognition. It ensures the high efficiency of calculation and makes dense events to be better expressed and information preserved.

Extensive experiments on multiple event-based classification datasets fully validated the effectiveness of our model. We achieve $99.1\%$, $99.7\%$, $99.7\%$ on the N-MNIST~\cite{orchard2015nmnist}, DVS128-Gait-Day~\cite{wang2021eventgait3Dgraph}, and ASL-DVS~\cite{bi2020graph} dataset, respectively.

The rest of this paper is organized as follows: 
In section~\ref{sec:relatedworks}, we review the related works on event-based recognition and graph neural networks. In section~\ref{sec:method}, we introduce our proposed method with a focus on the overview of our method, initial event representation, absorbing graph representation learning, classification head, and network training. Then, we conduct extensive experiments in section~\ref{sec:experiments}, and summarize our paper in section~\ref{sec:conclusion}.

\section{Related Work} \label{sec:relatedworks}

In this section, we give a review of Event-based Recognition and Graph Neural Networks. More related works can be found in the  surveys~\cite{gu2021HARsurvey, gallegoevent, wu2020GNNsurvey, yuan2022GNNtaxonomicSurvey} and paper list\footnote{\url{https://github.com/Event-AHU/Event_Camera_in_Top_Conference}}.

\noindent 
\textbf{Event-based Recognition.~} 
Current works can be divided into three streams for the event-based recognition, including the CNN based~\cite{wang2019evGait}, SNN based~\cite{fang2021PLIF, fang2021SNNIIR}, GNN based models~\cite{wang2021eventGNN, bi2019gnnevent, bi2020graph}, due to the flexible representation of event stream. 
For the CNN based models, Wang et al.~\cite{wang2019evGait} propose to identify human gaits using event camera and design a CNN model for recognition. 
SNN is also adopted to encode the event stream for energy-efficient recognition. For example,  
Peter et al.~\cite{diehl2015snnbalancing} propose the weight and threshold balancing method to achieve efficient ANN-to-SNN conversion. 
SNN-IIR~\cite{fang2021SNNIIR} is proposed by Fang et al. to search for the optimal synapse filter kernels and weights for SNN to learn the spatio-temporal patterns. 
Nicolas et al.~\cite{perez2021sparse} propose a sparse backpropagation method for SNNs that is faster and more memory efficient. 
Chen et al.~\cite{chen2022ecsnet} propose the compact event representation, termed 2D-1T event cloud sequence (2D-1T ECS), to exploit the inherent sparsity with reconciling the spatio-temporal information. A lightweight spatio-temporal learning framework (ECSNet) is proposed based on such event representation for object classification and action recognition. 
Deng et al.~\cite{deng2021mvfnet} project event stream to multi-view 2D maps and exploit spatio-temporal complements for event based tasks.

For point cloud based representation, Wang et al.~\cite{wang2019spaceCloud} treat the event stream as space-time event clouds and adopt PointNet~\cite{qi2017pointnet} as the backbone for gesture recognition. 
Sai et al. propose the event variational auto-encoder (eVAE)~\cite{vemprala2021representation} to achieve compact representation learning from the asynchronous event points directly. 
Fang et al.~\cite{fang2021snnresnet} propose SEW (spike-element-wise) residual learning for deep SNNs which addresses the vanishing/exploding gradient problems effectively. 
Meng et al.~\cite{meng2022DSR} propose an accurate and low latency SNN based on Differentiation on Spike Representation (DSR) method. 
TORE~\cite{baldwin2022TORE} is short for Time-Ordered Recent Event (TORE) volumes, which compactly stores the raw spike timing information. 
VMV-GCN~\cite{xie2022vmv} is proposed by Xie et al. which is a voxel-wise graph learning model to fuse multi-view volumetric. 
Li et al.~\cite{li2022eventFormer} introduce the Transformer network to learn event-based representation in a native vectorized tensor way. 
Different from these works, in this paper, we design a novel graph neural network to model the sampled voxels and points which can represent the event data 
more effectively via the specific message propagation mechanism.

\noindent 
\textbf{Graph Neural Networks.~} 
Due to the flexible formulation of graph, GNNs have been widely used in the computer vision community~\cite{wang2020learning, wang2021dualgnn, xiao2023dualGNN}. Some researchers also adopt GNNs for event data recognition. For example, the 3D graph neural network is proposed for gait recognition in work~\cite{wang2021eventGNN}. Bi et al.~\cite{bi2019gnnevent, bi2020graph} exploit the spatial and temporal feature learning for event based pattern recognition by employing residual-graph convolutional neural networks (RG-CNN) and Graph2Grid block. AEGNN~\cite{schaefer2022aegnn} (Asynchronous, Event-based Graph Neural Networks) is proposed to process events as evolving spatio-temporal graphs. Li et al.~\cite{LiYijin_2021_ICCV} propose  SlideGCN which achieves fast graph construction based on a radius search algorithm and rapid object recognition. 
Different from these works, in this paper, we propose to represent the event by designing novel absorbing point-voxel graphs. Then, we derive a specific absorbing GNN to effectively learn the representations for them. Finally, we integrate the representation cues of both point and voxel graphs for event-based recognition.

\section{Methodology} \label{sec:method}

In this section, we will first give an overview of our proposed model and initial event representation. Then, we will dive into the details of our absorbing graph representation learning, with a focus on graph construction, and absorbing graph convolutional networks (AGCN). After that, we introduce the classification head, and loss functions used in the training phase.

\subsection{Overview}  
\begin{figure*}
\center
\includegraphics[width=6.5in]{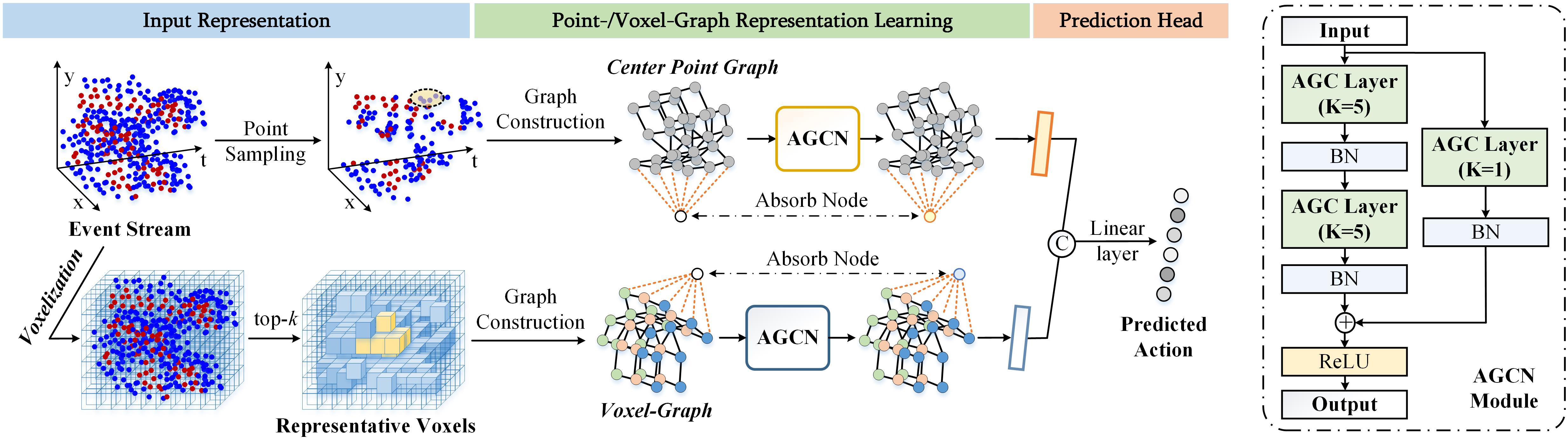}
\caption{ \textbf{An overview of our proposed Absorbing Graph Representation Learning framework for event-based recognition.} 
Specifically, we first transform the event stream into dual representations, i.e., the sparse event cloud and voxels. Then, we build dual graphs on the two inputs by treating each point/voxel-grid as the graph node. More importantly, we introduce absorbing nodes into the graph for the global information modeling.  The absorbing graph convolutional network (AGCN) is designed for structured feature learning and global feature aggregation, simultaneously. The outputs of dual branches will be concatenated and fed into a linear layer for final prediction.} 
\label{framework}
\end{figure*}

Given the input event stream with hundreds of thousands of events, we first adopt OctreeGrid filtering algorithm~\cite{lee2001point} and Voxel construction techniques to obtain (center) point and voxel  representations respectively. 
Then, we leverage two absorbing graphs, termed center point graph (CPoint-graph) and voxel graph (Voxel-graph) to model the 
spatio-temporal relationships of center  points and voxels respectively. 
After that, we devise a novel  absorbing graph convolutional network (AGCN) to learn effective feature descriptors for center and voxel graph based event representation.  
Finally, we aggregate them together for final event data  representation and  recognition. The overall framework is illustrated in Fig.~\ref{framework}. 
Below, we introduce the above modules in detail.

\begin{figure}
\center
\includegraphics[width=4in]{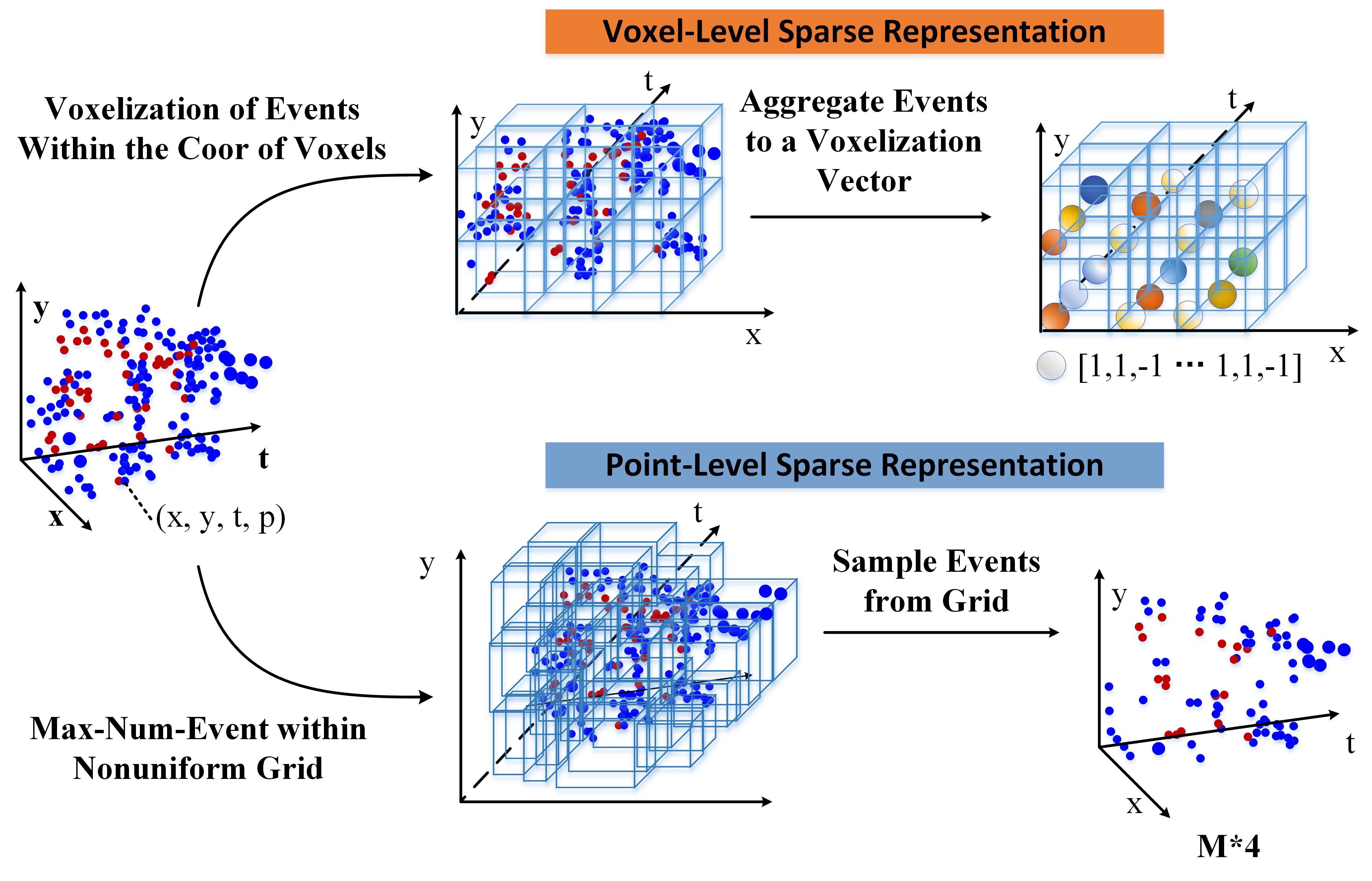}
\caption{ Illustration of our sparse point and voxel level down-sampling.}  
\label{down_sample}
\end{figure}

\subsection{Initial Event Representation} 
Considering the large amount of data and computational complexity, it is necessary to employ some down-sampling  techniques to reduce the number of events. In this paper, we adopt two kinds of sampling techniques to obtain the compressed event representations. To be specific, given original event stream $\mathcal{E}$ with $N$ events, we first apply the OctreeGrid filtering algorithm~\cite{lee2001point} to obtain $M$ 
representative events, denoted as \textbf{center points} $\mathcal{C}=\{c_1, c_2 \cdots c_M\}$. Each event point $c_i$ is expressed as a 4D tuple as
$$
c_i = (x_i,y_i,t_i,p_i)
$$
where $x_i, y_i$ denotes the spatial position and $t_i$ denotes the timestamp. The last $p_i$ represents the attribute/polarity. 
In our paper, we mainly focus on $(x_i,y_i,t_i)$ which represents the spatio-temporal coordinate/position of an event. In contrast to original events $\mathcal{E}$, the sampled $\mathcal{C}$ contains obviously fewer events and also can preserve the main spatio-temporal structure of the events.

In addition to center points  $\mathcal{C}$, we also employ voxelization to obtain 
voxel representations. Specifically, given the original event stream $\mathcal{E}$ with range $H, W, T$, we divide the spatio-temporal 3D space into voxels with the size of each voxel being  $h', w', t'$, as shown in Fig.~\ref{down_sample}. 
Hence,  each voxel generally contains several events and the resulting event voxels in spatio-temporal space are of size $H/h', W/w', T/t'$. 
In practice, the above voxelization 
usually still produces tens of thousands of voxels. 
In order to further reduce the number of voxels and alleviate the effect of noisy voxels, we also adopt a voxel selection process to select top $K$ voxels based on the number of events contained in each voxel. 
Let $\mathcal{O}=\{o_1, o_2 \cdots o_K\}$ denote the collection of the final selected voxels. 
Each event voxel  $o_i$ is associated with a feature descriptor $\textbf{a}_i\in \mathbb{R}^C$ which integrates  the attributes (polarity) of its involved events. 
Hence,  each $o_i\in \mathcal{O}$ is represented 
as 
$$
o_i=(x_i,y_i,t_i, \textbf{a}_i)
$$
where $x_i,y_i,t_i$ denotes the 
3D coordinate of each voxel.

\subsection{Absorbing Graph Representation Learning} 

After the above initial event representation $\mathcal{C}$ and $\mathcal{O}$, 
we then propose an effective  approach to learning discriminative representation for recognition tasks. 
As we know, the core feature encoded in the above $\mathcal{C}$ and $\mathcal{O}$ is the spatiotemporal relationship between event units (points or voxels). 
This motivates us to leverage the graph model and learning approach to represent the above pre-processed event streams respectively. 
In the following, we first introduce our graph construction for data $\mathcal{C}$ and $\mathcal{O}$. Then, we devise a new absorbing graph convolutional network (AGCN) to learn effective representations for event data $\mathcal{C}$ and $\mathcal{O}$.

\subsubsection{Graph Construction} 

In this work, two graphs are constructed for the point and voxel representations respectively. 

\noindent
\textbf{Center Points Graph. }  
For the center point event data $\mathcal{C}$, we construct a geometric neighboring graph $G^c(V^c, E^c)$. Specifically, 
each node $v_i\in V^c$ represents an event $c_i=(x_i,y_i,t_i,p_i)\in \mathcal{C}$ who has an attribute (polarity) $p_i \in \{1,-1\}$. 
An edge $e_{ij}\in E^c$  connects node $v_i$ and $v_j$ if their  distance is less than a threshold $R$, i.e., 
$$
d(c_i,c_j) <R
$$
where $R$ is pre-defined parameter. 
In our experiments, we define 
$d(c_i,c_j)$ as 
\begin{equation}
d(c_i,c_j) =\sqrt{(x_i-x_j)^2+(y_i-y_j)^2+(t_i-t_j)^2}
\end{equation}
In particular, an additional absorbing node $\tilde{v}$ is added to $V^c$. It is connected to all event nodes, i.e., there exists an edge between $\tilde{v}$ and every event node $v_i$ where $i=1,2\cdots M$. 
Fig.~\ref{framework} shows the construction of the proposed  Center Point Graph. 
Note that, the main purpose of node $\tilde{v}$ is to absorb/integrate  the representation messages from all event nodes to obtain the global representation for the whole graph, as discussed in Section~\ref{impleDetails}.

\noindent 
\textbf{Voxel Graph. } 
For  voxel event data $\mathcal{O}$, we similarly construct a geometric neighboring graph $G^o(V^o,E^o)$. To be specific,  
each node $v_i\in V^o$ represents a voxel $o_i=(x_i,y_i,t_i,\textbf{a}_i)\in \mathcal{O}$ which is described as a feature vector $\textbf{a}_i \in \mathbb{R}^{C}$. 
The edge $e_{ij}\in E^o$ exists between node $v_i$ and $v_j$ if the Euclidean distance 
between their 3D coordinates is less than a threshold $R_2$, as shown in Eq.(1). 
Also,  an absorbing node ${v'}$ which connects to all voxel nodes  is also added to $V^o$. It aims to absorb/integrate the information from all event nodes to obtain the global-level representation for the whole voxel graph. Fig.~\ref{framework} shows the illustration of the voxel graph construction.

\subsubsection{Absorbing GCN}

Based on the above center point  and voxel graphs with absorbing nodes, we then propose a novel Absorbing GCN (AGCN) model to learn the effective  representations for them respectively.  
The proposed AGCN consists of several learning layers with residual connection between the first and last layers, as shown in Fig.~\ref{framework} (right). 
Each layer conducts message passing on the graph. 
To be specific, in each AGCN layer, each event node $v_i$  aggregates the features from its adjacency nodes as 
\begin{align}
f'_d(v_i)\leftarrow \sigma\Big(\sum_{v\in\{\mathcal{N}(v_i)\bigcup \tilde{v}\}}\omega_{d}(v_i,v) f(v)\Big), d = 1, 2\cdots D
\tag{2}
\end{align}
Also, the absorbing node $\tilde{v}$ aggregates the messages from all other nodes as
\begin{align}
&f'_d(\tilde{v})\leftarrow \sigma\Big(\sum_{v\in V}\omega_{d}(\tilde{v},v) f(v)\Big), d = 1, 2\cdots D
\tag{3}
\end{align}
where  $\sigma(\cdot)$ denotes the activation function, such as ReLU. $\omega_{d}(v_i,v)$ and $\omega_{d}(\tilde{v},v)$ denote the learnable convolution kernel weights. 
Following works~\cite{monti2017geometric,wang2021eventgait3Dgraph},
 we define them as Gaussian Mixture
Model (GMM) function~\cite{monti2017geometric} on the pseudo-coordinate. 
Generally, given any node pair $(u,v)$, we first calculate its 
pseudo-coordinate~\footnote{For each node pair $(u,v)$ with ${z}_{u,v} = (\frac{1}{\sqrt{deg(u)}},\frac{1}{\sqrt{deg(v)}}) $} as ${z}_{u,v}$. 
Then, we learn the weight kernel $\omega_{d}(u,v)$
as 
\begin{equation}
\omega_{d}(u,v)= \\
\sum^K_{k=1} \alpha_k
    \exp(-\frac{1}{2}(z_{uv}-\mu_k)^T{\Sigma_k }^{-1}(z_{uv}-\mu_k))\tag{4}
\end{equation}
where $\mu_k, \Sigma_k$ are learnable parameters and $\alpha_k$ denotes the weight of the $k$-th Gaussian kernel. $K$ denotes the number of Gaussian kernels. 

Using the above layer-wise message passing, we can define multi-layer AGCN  architecture with the residual connection between the first and last layers, as shown in Fig.~\ref{framework} (right).
We use $Y^c\in\mathbb{R}^{M\times d}$, $ Y^o\in \mathbb{R}^{L\times d}$ to denote the outputs of two branches after using AGCN module, i.e., 
\begin{align}
 Y^c = \mathrm{AGCN} (G^c, \Omega^c),\,\, Y^o = \mathrm{AGCN} (G^o, \Omega^o)\tag{5}
\end{align}
where $\Omega^c$ and $\Omega^o$ denote the all parameters of two branches.

\subsection{Classification Head and Network Training}

Let $Y^c_{\tilde{v}}$ and $Y^o_{\tilde{v}}$ denote the representations for the absorbing nodes in both center point and voxel graphs respectively. 
As discussed in \S 3.3.2, since the absorbing node can aggregate the information from all event nodes, we can render its representation as the global graph-level representation. 
Therefore, we  concatenate  $Y^c_{\tilde{v}}$ and $Y^o_{\tilde{v}}$ together and adopt MLP to predict the final class label as
\begin{equation}
 Y = \mathrm{MLP} (Y^c_{\tilde{v}} \|Y^o_{\tilde{v}})\tag{6}
\end{equation}
where $\|$ denotes the concatenation operation.  
In addition, we add some dropout and batch-normalization layers between MLP layers to avoid the possible issues of over-fitting and gradient extinction. 
The whole network is trained in an end-to-end manner. 
We adopt the Negative Log Likelihood Loss~\cite{miranda2017understanding} as our loss function to train the whole network. 


\subsection{Discussion on Feature Aggregation Module}   
Existing neural networks usually adopt the \textit{pooling layer} and \textit{CLS token} to achieve feature aggregation.  The obtained features are used as the input for subsequent layers or used for the head network of each downstream task. Obviously, our proposed absorbing node is significantly different from existing ones: 
\textit{1). Different Application Scenarios:~} 
The pooling layers are widely used in convolutional neural networks or recurrent neural networks and the CLS token is designed for Transformer networks. In contrast, our proposed absorbing node is specifically proposed for information aggregation in the graph neural networks. 
\textit{2). Different Working Principles:~} 
The pooling layer directly selects the maximum or average values into one representative value, corresponding to the max-pooling or average-pooling operation. The CLS token is usually randomly initialized as the input of Transformer networks and updates its values with the training of the network. The proposed absorbing node is embedded in the graph structure for deep feature interaction and information aggregation. It can be used as a bridge of information transmission, thus aggregating important information from other nodes, which helps to improve the execution efficiency of subsequent head networks of tasks.

\section{Experiments}  \label{sec:experiments}


\subsection{Dataset and Evaluation Metric}  \label{datasetMetric}

In this paper, our experiments are conducted on four event-based classification datasets, including \textbf{ASL-DVS}~\cite{bi2020graph}, \textbf{N-MNIST}~\cite{orchard2015nmnist}, \textbf{DVS128-Gait-Day}~\cite{wang2021eventgait3Dgraph}, and \textbf{HARDVS}~\cite{wang2022hardvs} dataset. A brief introduction to these datasets is given below. 
The \textbf{top-1 accuracy} is adopted as the evaluation metric for the evaluation of our proposed model and other SOTA recognition approaches.  

$\bullet$ \textbf{ASL-DVS}~\cite{bi2020graph} is a large 24-class event-based recognition dataset proposed by Bi et al. in the year 2020. It focuses on the handshape corresponding to 24 letters from the American Sign Language recorded under realistic conditions using an iniLabs DAVIS240c camera. Some representative samples. The authors record 100,800 samples for this dataset (4,200 samples for each letter), and each video lasts for about 100 milliseconds. In our experiments, for each category, we randomly split the videos into training and testing subset which contains $80\%$ (80,640) and $20\%$ (20,160), respectively. 

$\bullet$ \textbf{DVS128-Gait-Day}~\cite{wang2021eventgait3Dgraph} dataset is proposed for event-based gait recognition. It contains 4,000 videos corresponding to 20 classes. 20 volunteers are recruited for data collection using a DVS128 Dynamic Vision Sensor (the pixel resolution is $128 \times 128$). 

$\bullet$ \textbf{N-MNIST}~\cite{orchard2015nmnist} dataset is obtained by recording the display equipment when visualizing the original MNIST (28 × 28 pixels). The ATIS event camera is used for the data collection and each event sample  lasts about 10ms. There are 70,000 event files for this dataset, the training and testing subset contains 60,000 and 10,000 videos, respectively.

$\bullet$ \textbf{HARDVS}~\cite{wang2022hardvs} is a newly released large-scale event-based human activity recognition dataset in the year 2022. It contains 107,646 aligned RGB and event modalities corresponding to 300 wide human activities recorded using a DVS346 event camera. The spatial resolution is $346 \times 260$ and each video lasts for about 5 seconds. Various challenges are considered in this dataset, such as multi-view, illumination, motion, dynamic background, occlusion, etc. Following~\cite{wang2022hardvs}, we adopt 64,526 and 32,386 videos as training and testing subsets. More details can be found on the GitHub\footnote {\url{https://github.com/Event-AHU/HARDVS}}.



\subsection{Implementation Details} \label{impleDetails}
When conducting the non-uniform grid down-sampling, the MaxNumEvents is set as 40 for the DVS128-Gait-Day and ASL-DVS datasets, and 6,120 for N-MNIST and HARDVS datasets. The threshold $R$ is set as 5 for the point-based graph construction. We normalize $t$ to a preset threshold to make it consistent with the spatial range. Different voxel grids are selected for various datasets, more in detail, 2048, 512, 512, 512 are chosen for the DVS128-Gait-Day, ASL-DVS, N-MNIST and HARDVS datasets. After considering the spatio-temporal discrepancy across different datasets, we set the scale ($v_{h},v_{w},v_{t}$) of voxel grid as (10, 10, 10) for ASL-DVS, (4, 4, 4) for DVS128-Gait-Day, (20, 2, 2), (50, 30, 20) for  N-MNIST and HARDVS datasets. When building graphs for the voxel branch, the threshold $R$ is set as 2. 

In the training phase, our model is trained for 150 epochs with a learning rate of 0.001 and decaying the learning rate by a factor of 0.1 on the $60^{th}$ and $110^{th}$ epoch. The Adam~\cite{kingma2014adam} is selected as our optimizer. Note that, all these important hyperparameters are shared for the four datasets in our experiments. Our code is implemented based on Python and PyTorch~\cite{paszke2019pytorch} deep learning framework. The experiments are conducted on a server with RTX3090 GPUs.

\subsection{Comparison with Other SOTA Algorithms} \label{benchmarkResults}

In this section, we will report and compare our results with other state-of-the-art performances on four event-based classification datasets. 

\noindent
\textbf{Results on ASL-DVS~\cite{bi2020graph}. } 
As shown in Table~\ref{ASLDVSResults}, we can find that our baseline method Ev-Gait-3DGraph~\cite{wang2021eventgait3Dgraph} achieves 0.738 on the ASL-DVS dataset. In contrast, our proposed framework achieves a significant gain of $+0.259$, i.e., 0.997 on the top-1 accuracy. To the best of our knowledge, it is a new state-of-the-art on this benchmark dataset. More in detail, we can find that the compared methods EST~\cite{gehrig2019end}, M-LSTM~\cite{cannici2020mlstm}, and EventNet~\cite{sekikawa2019eventnet} achieve 0.979, 0.980, and 0.833, which are all inferior to ours. Our results are also better than the CNN-based models (such as ResNet-50~\cite{he2016resnet} and RG-CNNs~\cite{bi2020graph}) and GNN-based models (EV-GCNNs~\cite{deng2021evvgcnn} and VMV-GCN~\cite{xie2022vmvgcn}). These experimental results fully validated the effectiveness of our proposed framework.  

\begin{table}[!htp]
\center
\small    
\caption{Results on the ASL-DVS~\cite{bi2020graph} dataset.} 
\label{ASLDVSResults}
\scalebox{0.8}{
\begin{tabular}{ccccccccccccccc} 		
\hline \toprule [0.5 pt] 
\textbf{EST}\cite{gehrig2019end}   &\textbf{AMAE}\cite{deng2020amae}     &\textbf{M-LSTM}\cite{cannici2020mlstm}    &\textbf{MVF-Net}\cite{deng2021mvfnet}     & \textbf{EventNet}\cite{sekikawa2019eventnet}\\  
0.979   & 0.984     &0.980     &0.971     &0.833    \\ 
\hline 
\textbf{RG-CNNs}\cite{bi2020graph}     &\textbf{\makecell[c]{EV-VGCNN}}\cite{deng2021evvgcnn}     &\textbf{\makecell[c]{VMV-GCN}}\cite{xie2022vmvgcn}     &\textbf{EV-Gait-3DGraph}\cite{wang2019evGait}   &\textbf{Ours} \\
0.901     &0.983     &0.989  &0.738  &0.997	 \\
\hline \toprule [0.5 pt] 
\end{tabular}
}
\end{table}


\begin{table*}
\center
\scriptsize    
\caption{Results on the N-MNIST~\cite{orchard2015nmnist} Dataset.} 
\label{N-MNISTResults} 
\scalebox{0.9}{
\begin{tabular}{ccccccccccccccccc} 		
\hline \toprule [0.5 pt] 
\textbf{\makecell[c]{EST}}\cite{gehrig2019end}   &\textbf{\makecell[c]{M-LSTM}}\cite{cannici2020mlstm} 
&\textbf{\makecell[c]{MVF-Net}}\cite{deng2021mvfnet}  &\textbf{\makecell[c]{Gabor-SNN}}\cite{sironi2018hats} 
&\textbf{EvS-S}\cite{li2021graph} &\textbf{HATS}\cite{sironi2018hats}     &\textbf{EventNet}\cite{sekikawa2019eventnet}     &\textbf{RG-CNNs}\cite{bi2020graph}   &\textbf{EV-VGCNN}\cite{deng2021evvgcnn} &\textbf{EV-Gait-3DGraph}\cite{wang2021eventgait3Dgraph} &\textbf{Ours}  \\ 
99.0   &98.6     &98.1     &83.7     &99.1  &99.1   &75.2     &99.0     & 99.4	&96.6  &99.1	\\ 
\hline \toprule [0.5 pt] 
\end{tabular}
}
\end{table*}

\begin{table} 
\centering 
\caption{Results on the HARDVS~\cite{wang2022hardvs} dataset. w/o denotes without the following operation.}   
\label{hardvsResults}
\small 
\begin{tabular}{c|c|c|c|c|c|c|c} 
\hline \toprule [0.5 pt] 
&\textbf{No.} &\textbf{Algorithm} &\textbf{Publish}  &\textbf{Backbone} &\textbf{FLOPS} &\textbf{Param}  &\textbf{Top1} \\
\hline
\multirow{8}{*}{\rotatebox{90}{With Pre-train}}
&01 & ResNet18~\cite{he2016resnet} &CVPR-2016    &ResNet18   &8.6G	&11.7M  &49.20      \\
\cline{2-8}
&02 & C3D~\cite{tran2015c3d} &ICCV-2015   &CNN  &0.1G	&147.2M &50.52    \\
\cline{2-8}
&03 & TimeSformer \cite{bertasius2021TimeSformer}  &ICML-2021     &VIT    &53.6G	&121.2M  &50.77     \\
\cline{2-8}
&04 & TSM ~\cite{lin2019tsm} &ICCV-2019     &ResNet-50    &0.3G	&24.3M   &52.63     \\
\cline{2-8}
&05 & ACTION-Net~\cite{wang2021actionnet} &CVPR-2021     &ResNet-50   &17.3G	&27.9M    &46.85   \\
\cline{2-8}
&06 & TAM \cite{liu2021tam}  &ICCV-2021     &ResNet-50   &16.6G	&25.6M    &50.41     \\
\cline{2-8}
&07 &  Video-SwinTrans \cite{liu2021videoSwin}   &CVPR-2022     &Swin-T.F.   &8.7G	&27.8M   &51.91     \\
\cline{2-8}
&08 & ESTF\cite{wang2022hardvs}  &-       &ResNet18     &8.8G	&27.8M  &51.22     \\
\hline
\multirow{5}*{\rotatebox{90}{w/o Pretrain}}
&09 & SlowFast  \cite{feichtenhofer2019slowfast}  &ICCV-2019     &ResNet-50    &0.3G	&33.6M   &46.54    \\
\cline{2-8}
&10 & X3D  \cite{feichtenhofer2020x3d}  &CVPR-2020     &ResNet     &0.9G	&2.1M  &45.82    \\
\cline{2-8}
&11 & R2Plus1D \cite{tran2018R2Plus1D} &CVPR-2018 &ResNet-18   &20.3G	&63.5M   &49.06    \\
\cline{2-8}
&12 & EV-Gait-3DGraph \cite{wang2021eventgait3Dgraph}  &T-PAMI-2021   &-   &-	&-    &23.20     \\
\cline{2-8}
&13 & Ours  &-       &-    &23.3G	&7.1M   &49.50  \\
\hline \toprule [0.5 pt] 
\end{tabular}
\end{table}

\noindent
\textbf{Results on DVS128-Gait-Day~\cite{wang2021eventgait3Dgraph}. } 
As shown in Table~\ref{DVS128Caltech101Results}, our proposed model achieves $99.7\%$ on the top-1 accuracy metric which is significantly better than all the compared methods. To be specific, our model outperforms the baseline Ev-Gait-3DGraph~\cite{wang2021eventgait3Dgraph} by $+4.8\%$. This comparison fully demonstrates the effectiveness of our proposed joint point-voxel event representation and absorbing nodes used in graph neural networks. A similar conclusion can also be drawn from the comparison with other methods including 2DGraph-3DCNN~\cite{bi2020graph} ($92.2\%$), EV-Gait-IMG~\cite{wang2021eventgait3Dgraph} ($87.3\%$), LSTM-CNN~\cite{wang2021eventgait3Dgraph} ($86.5\%$). 

\begin{table}[!htp]
\center
\small   
\caption{Results on the DVS128-Gait-Day~\cite{wang2021eventgait3Dgraph} dataset.} 
\label{DVS128Caltech101Results} 
\scalebox{0.9}{
\begin{tabular}{ccccccc} 		
\hline \toprule [0.5 pt] 
\textbf{\makecell[c]{EVGait\\-3DGraph\\\cite{wang2019evGait}}}   &\textbf{\makecell[c]{2DGraph\\-3DCNN\\\cite{bi2020graph}}} &\textbf{\makecell[c]{EV-Gait\\-IMG\\\cite{wang2019evGait}}} 
&\textbf{\makecell[c]{LSTM\\-CNN\\\cite{bi2020graph}}}   &\textbf{\makecell[c]{SVM\\-PCA\\\cite{cortes1995support}}}  &\textbf{Ours}  \\
94.9   &92.2     &87.3     &86.5     &78.05     &99.7     	  		 \\
\hline \toprule [0.5 pt] 
\end{tabular}
}
\end{table}

\noindent
\textbf{Results on N-MNIST~\cite{orchard2015nmnist}.}
From the experimental results reported in Table~\ref{N-MNISTResults}, it is easy to find that the baseline approach EV-Gait-3DGraph~\cite{wang2021eventgait3Dgraph} obtains $96.6\%$ on the N-MNIST dataset, while ours are $99.1\%$ is ranked the second place on this benchmark dataset. This result is comparable with recent strong event-based classification models, including EvS-S~\cite{li2021graph}, HATS~\cite{sironi2018hats}, RG-CNNs~\cite{bi2020graph}, and EV-VGCNN~\cite{deng2021evvgcnn}. Our model is also better than M-LSTM~\cite{cannici2020mlstm}, MVF-Net~\cite{deng2021mvfnet}, Gabor-SNN~\cite{sironi2018hats}, EventNet~\cite{sekikawa2019eventnet}. The effectiveness of joint dual event stream representation and absorbing graph neural networks are validated by these results.

\noindent
\textbf{Results on HARDVS~\cite{wang2022hardvs}.} 
As shown in Table~\ref{hardvsResults}, our model achieves $49.50\%$ on this large-scale dataset which is better than our baseline ($23.2\%$) by $+26.3\%$ on the top-1 accuracy. It is worth noting that part of the compared methods adopts pre-trained model weights on other tasks. For example, the pre-trained weights of ResNet-18~\cite{he2016resnet}, ResNet-50~\cite{he2016resnet}, and ViT~\cite{dosovitskiy2020image} on the ImageNet classification dataset are used in ESTF~\cite{wang2022hardvs}, TSM~\cite{lin2019tsm}, ACTION-Net~\cite{wang2021actionnet}, TAM~\cite{liu2021tam}, TimeSformer~\cite{bertasius2021TimeSformer}, etc. Nevertheless, we still achieve better results than the ResNet-18~\cite{he2016resnet}, and ACTION-Net~\cite{wang2021actionnet}. Compared with the methods trained from scratch, we beat the SlowFast~\cite{feichtenhofer2019slowfast} (46.54), X3D~\cite{feichtenhofer2020x3d} (45.82), and R2Plus1D~\cite{tran2018R2Plus1D} (49.06). These results fully demonstrate the effectiveness of our proposed modules for event-based classification.

\begin{table}
\center   
\caption{Ablation Study on the DVS128-Gait-Day Dataset.} \label{componentAnalysisResults} 
\begin{tabular}{c|ccc|ccc} 		
\hline \toprule [0.5 pt] 
\textbf{No.} & \textbf{Voxel} &\textbf{Points} &\textbf{Proxy-Node} &\textbf{Accuracy}  \\
\hline 
1 &\textcolor{DarkRed}{\xmark}     &\textcolor{SeaGreen4}{\cmark}     &\textcolor{DarkRed}{\xmark}          &94.9        	  		 \\
\hline 
2 &\textcolor{SeaGreen4}{\cmark}    &\textcolor{DarkRed}{\xmark}     &\textcolor{SeaGreen4}{\cmark}          &96.5         	  		 \\
\hline
3 &\textcolor{DarkRed}{\xmark}     &\textcolor{SeaGreen4}{\cmark}     &\textcolor{SeaGreen4}{\cmark}          &96.5 	  		 \\
\hline
4 &\textcolor{SeaGreen4}{\cmark}     &\textcolor{SeaGreen4}{\cmark}     &\textcolor{SeaGreen4}{\cmark}          &99.7        	  		 \\
\hline \toprule [0.5 pt] 
\end{tabular} 
\end{table}

\subsection{Ablation Study} \label{ablationStudy} 

In this section, we conduct experiments of component analysis and different settings to check their influence.

\noindent
\textbf{Effects of Dual-View Learning.~} 
In this work, we represent the event streams using dual views, i.e., the point cloud and voxel. We analyze the two kinds of event representation on the DVS128-Gait-Day dataset~\cite{wang2021eventgait3Dgraph} and check whether the dual views contribute a better event-based representation learning. As illustrated in Table~\ref{componentAnalysisResults}, the top-1 accuracy is all $96.5\%$ when only one modality is used. More importantly, our results can be improved to $99.7\%$ when the point cloud and voxel branch are used simultaneously. This experiment demonstrates that the joint representation indeed works for the event-based classification. This also validated that the trade-off between the accuracy and complexity of the model can be further improved.

To further validate the aforementioned conclusions, we also design two variants including both \emph{parallel} and \emph{stacked} GCNs that have the same parameters as our AGCN and compare them with our AGCN model on the DVS128-Gait-Day dataset. As illustrated in Table~\ref{doubleGNNmodels}, we can find that the overall performance can be slightly improved using parallel GCNs by comparing the algorithm \#1 and \#2, and algorithm \#4 and \#5. However, simply stacking the GCN layers will decrease the final recognition results by comparing algorithm \#1 and 3, and algorithm \#4 and \#6. When we combine the point and voxel event representations together, i.e., Double Point-Voxel-GCN (Parallel), the recognition performance can be significantly improved. These results fully validated the effectiveness of dual-representation learning for event-based classification. Note that, this result can be further improved when incorporating the absorbing node for classification, i.e., 99.7 on the top-1 accuracy.

\begin{table}
\center   
\caption{Compared with GNN models with double parameters (stacked and parallel version)} 
\label{doubleGNNmodels} 
\begin{tabular}{c|cccccc} 		
\hline \toprule [0.5 pt] 
\textbf{No.} & \textbf{Method} &\textbf{Accuracy}  \\
\hline 
1 &Vanilla Point-GCN	&94.9  \\ 
2 &Double Point-GCN (Parallel)	&94.9    \\ 
3 &Double Point-GCN (stacked)	&91.4 \\ 
\hline 
4 &Vanilla Voxel-GCN	&94.6 \\ 
5 &Double Voxel-GCN (Parallel)	&95.9 \\
6 &Double Voxel-GCN (stacked)	&92.7 \\
\hline 
7 &Double Point-Voxel-GCN (Parallel)	&97.3 \\
8 &Our AGCN	&99.7 \\ 
\hline \toprule [0.5 pt] 
\end{tabular} 
\end{table}

\begin{figure} 
\center
\includegraphics[width=4.5in]{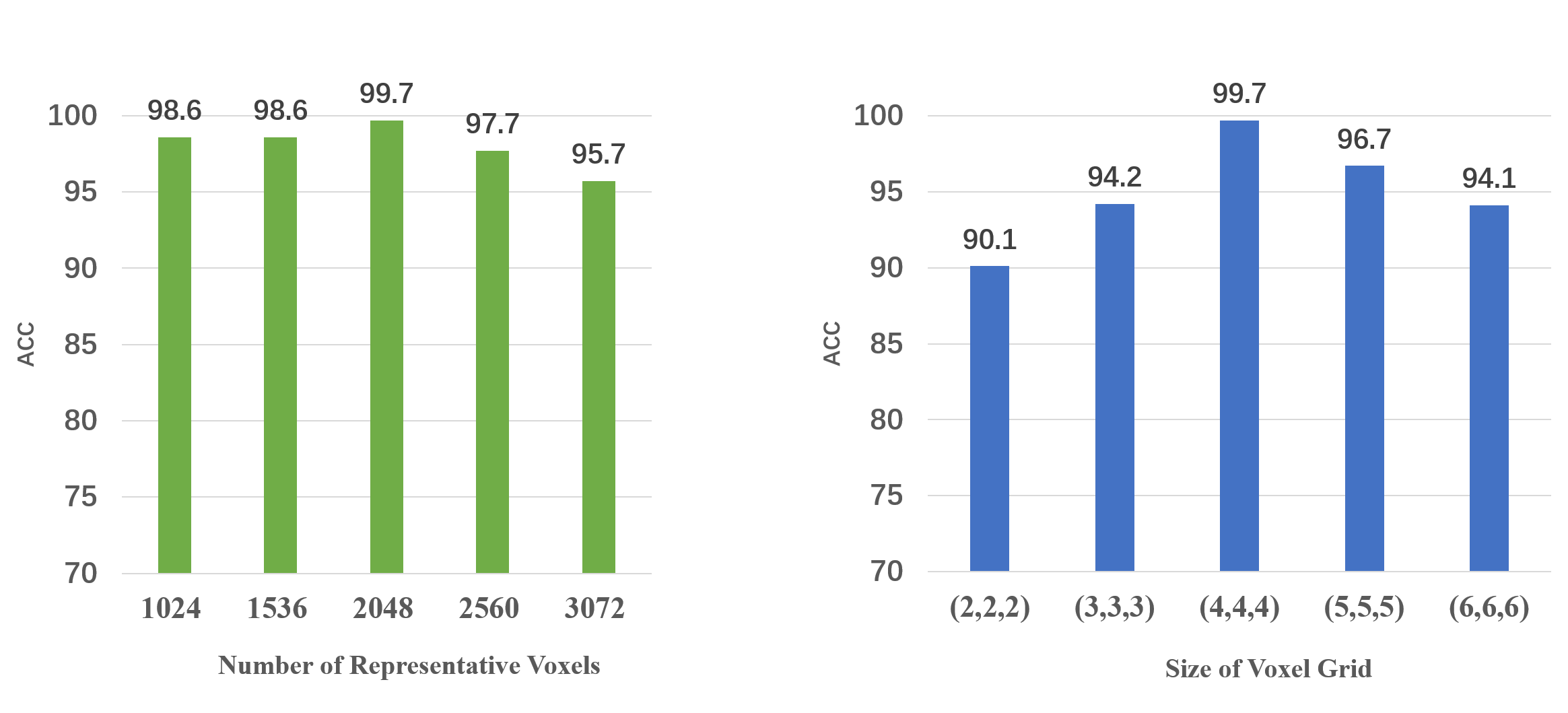}
\caption{Results of the different numbers of representative voxels and various scales of voxel-grid.} 
\label{PA_voxels}
\end{figure}

\noindent
\textbf{Effects of Absorbing Node.~} 
Different from previous works which usually separate the structured graph representation learning and information aggregation into two stages, in this work, we propose joint learning of these feature representations in a unified phase. The absorbing nodes are introduced for global information aggregation when building our point-/voxel-graph. As shown in Table~\ref{componentAnalysisResults}, we can find that the overall performance can be improved from $94.9\%$ to $96.5\%$ which demonstrates that our single-stage feature learning and aggregation works better than the previous two stage-based algorithms.

\begin{figure*} 
\center
\includegraphics[width=6.5in]{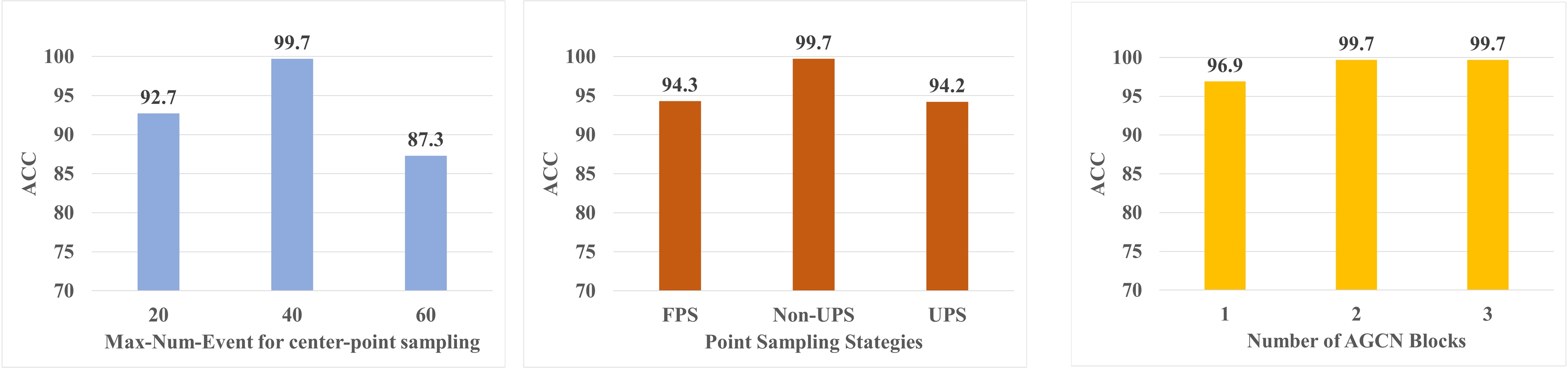}
\caption{Results of different Max-Num-Event for sampling, point sampling strategies and various numbers of AGCN blocks.} 
\label{PA_pointsampling}
\end{figure*}

\noindent
\textbf{Analysis on Number of Representative Voxels. } 
Representative voxel-grid sampling is a key step in our proposed framework. In this part, we test the various number of voxel-grids and check the influence of this parameter. As shown in the left part of Fig.~\ref{PA_voxels}, we select 1024, 1536, 2048, 2560, and 3072 voxel-grids to build our voxel-graph and obtain 98.6, 98.6, 99.7, 97.7, 95.7 on the top-1 accuracy metric respectively. Note that, the performance is firstly improved when increasing from 1024 to 2048, but dropped when a larger number of voxel-grids are selected. We think this may be caused by the fact that the less informative voxel-grids increase the difficulty of model computation. Therefore, we select 2048 voxel-grids in our other experiments.

\noindent
\textbf{Analysis on Size of Voxel Grid.~} 
The scale of voxel-grid is another key parameter in our proposed framework. As illustrated in the right part of Fig.~\ref{PA_voxels}, we set various sizes to check their final results, i.e., (2, 2, 2), (3, 3, 3), (4, 4, 4), (5, 5, 5), (6, 6, 6). It is easy to find that the best performance can be obtained when the scale is (4, 4, 4), i.e., 99.7 on the DVS128-Gait-Day dataset.

\noindent
\textbf{Analysis on Number of AGCN Blocks.~} 
The AGCN blocks can be adjusted based on the difficulty of the event-based recognition task. We tune this parameter on the DVS128-Gait-Day dataset and set it as 1, 2, and 3, respectively. We can find that the results are relatively stable when less than 3 blocks, i.e., 96.9 and 96.7 on top-1 accuracy. Better results can be obtained when three AGCN blocks are used, i.e., 99.7 on this dataset. Stacking more blocks didn't see any significant improvement.

\noindent
\textbf{Analysis on Max-Num-Event for Point sampling.~} 
The non-uniform sampling strategy is conducted on the dense event stream in our framework. More in detail, it randomly reserves one of the event points as the output. Therefore, the maximum number of events captured for each sampling operation (short for Max-Num-Event) is an important parameter for final results. In this part, we set different numbers as 20, 40, and 60 and their corresponding results are 92.7, 99.7, and 87.3 on the top-1 accuracy respectively. We can find that a better result can be obtained when this parameter is set to 40.

\noindent
\textbf{Analysis on Different Point Sampling Strategies.~}  
In this subsection, we conduct an ablation study on the different point sampling strategies, including the widely used Farthest Point Sampling (FPS), Uniform Point Sampling (UPS), and Non-Uniform Point Sampling (Non-UPS) strategy. As shown in Fig.~\ref{PA_pointsampling}, the FPS-based method achieves 94.3, the UPS-based model achieves 94.2, and the Non-UPS-based model obtains significantly better results, i.e., 99.7 on the DVS128-Gait-Day dataset. We can find that the Non-Uniform Point Sampling (Non-UPS) strategy will be more suitable for the event-based classification task, as it preserves better spatiotemporal information.

\subsection{Model Parameters and Efficiency Analysis} \label{paraeffAnalysis}
The size of the saved model of our baseline method EV-Gait-3DGraph~\cite{wang2021eventgait3Dgraph} is 29.3 MB, while ours is 26.9 MB. For the running efficiency, the baseline is 0.015 seconds per sample, meanwhile, ours is 0.02. We also report the FLOPS and parameters of the compared methods and ours in Table~\ref{hardvsResults}, we can find that our model is relatively small-scale meanwhile achieving high recognition performance.

\begin{figure*} 
\center
\includegraphics[width=6in]{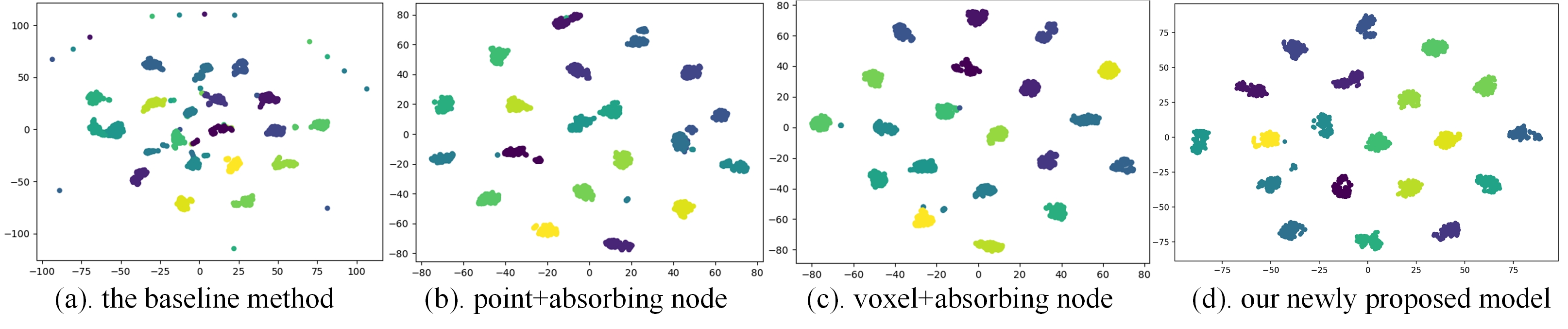}
\caption{Visualization of feature distribution on the DVS128-Gait-Day dataset.}  
\label{featureVIS_DVS128gaitday}
\end{figure*}

\begin{figure*} 
\center
\includegraphics[width=6in]{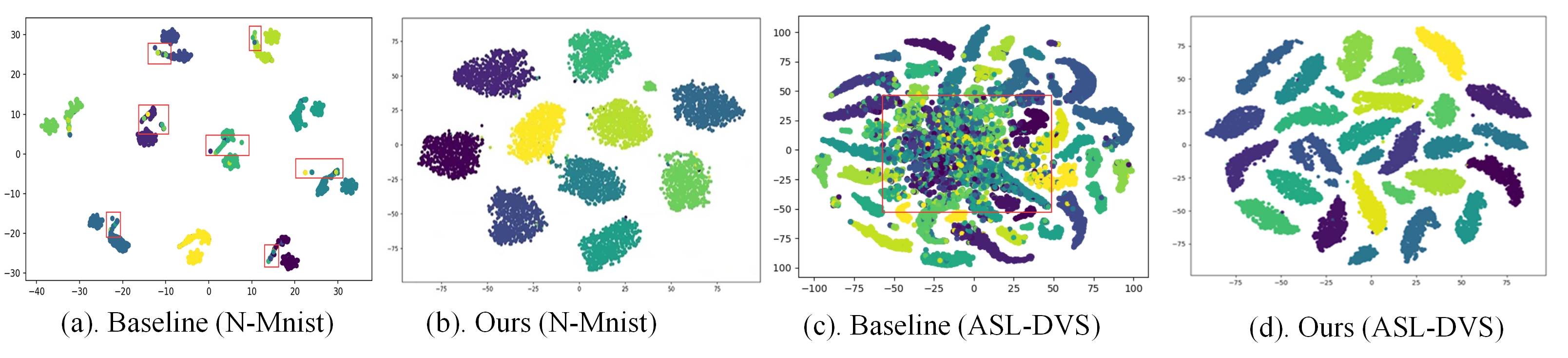}
\caption{Visualization of feature distribution on the N-MNIST dataset and ASL-DVS dataset. (a, c) Baseline approach, (b, d) Ours.}  
\label{featureVIS_ASLDVS_NMnist}
\end{figure*}

\begin{figure*}
\center
\includegraphics[width=6in]{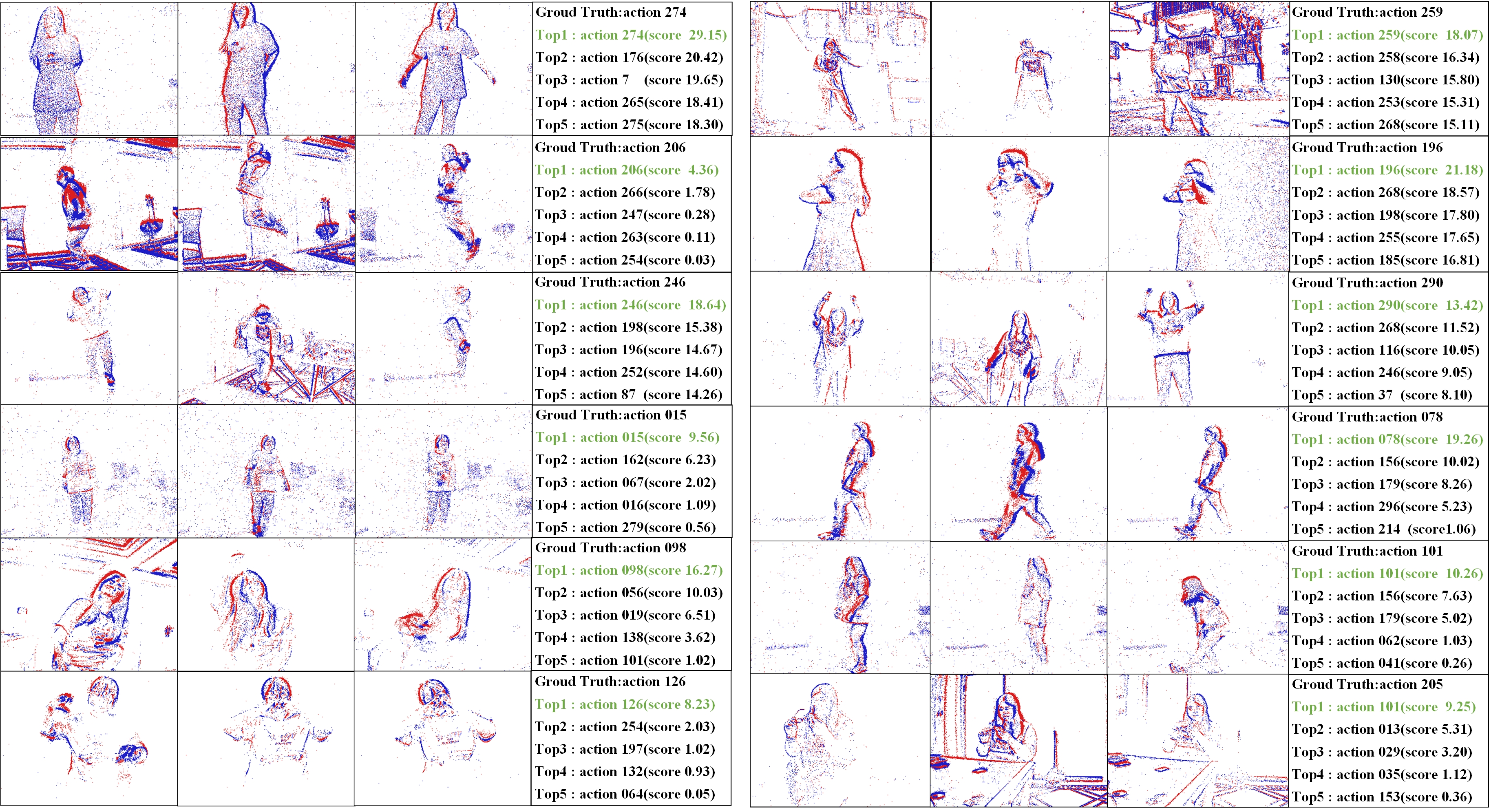}
\caption{Visualization of the top-5 predicted actions using our model on the HARDVS dataset.} 
\label{top5_hardvs}
\end{figure*}

\begin{figure*}
\center
\includegraphics[width=6in]{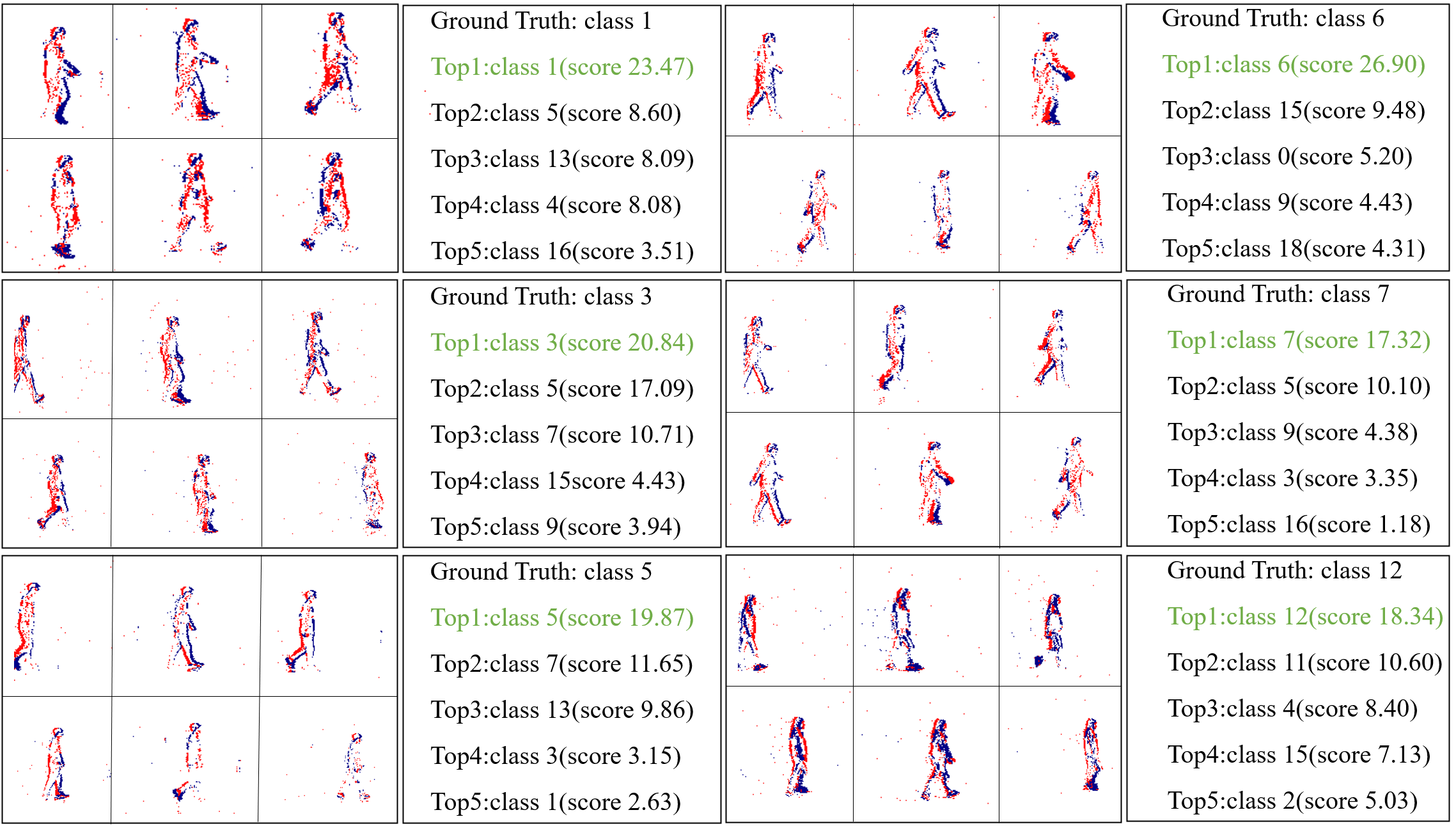}
\caption{Visualization of the top-5 predicted actions using our model on the DVS128-Gait-Day dataset.} 
\label{top5}
\end{figure*}

\subsection{Visualization}  \label{Visualization}

In addition to the aforementioned quantitative analysis, in this paper, we also give the qualitative analysis to help the readers better understand the effectiveness of our model. The structured features are visualized using the TSNE toolkit \footnote{\url{https://github.com/mxl1990/tsne-pytorch}} and also the recognition results are all provided in the following paragraphs, respectively.

\noindent
\textbf{Visualization of Feature Embedding.~} 
Different from regular CNN-based models, we can't visualize the feature maps from the spatial view. In this part, we adopt the TSNE toolkit to give a visualization of the learned structured feature representation. As shown in Fig.~\ref{featureVIS_DVS128gaitday}, we provide four groups of visualized features to compare, including (a) our baseline approach, (b) point+absorbing, (c) voxel + absorbing node, and (d). our newly proposed model. It is easy to find that our baseline is inferior to the methods after integrating our newly proposed modules (b, c, d). The feature distance between different categories is further, and the inside of the same cluster is more compact. Similar views can also be drawn from the feature visualization of the other two datasets (Fig.~\ref{featureVIS_ASLDVS_NMnist}), i.e., the N-MNIST and ASL-DVS dataset. Therefore, we can draw the conclusion that our proposed modules are effective for the event-based classification task.

\noindent
\textbf{Visualization of Recognition Results.~} 
As shown in Fig.~\ref{top5}, we give some visualizations of our proposed method on the DVS128-Gait-Day dataset. We can find that our model can predict the right category with a relatively higher response score.

\subsection{Limitation Analysis}  \label{sec::limitAnaly}  

Thanks to our proposed joint point-voxel based event representation and AGCN network, we achieve better performance on multiple benchmark datasets. Actually, our model can still be improved from the following aspects: 
1). Our proposed AGCN model is trained from scratch which can't make full use of the pre-trained weights on other tasks. Although good performance can be obtained, however, the current version still fails to transfer knowledge from large-scale data. 
2). We validate our proposed model on multiple event datasets with low and moderate resolutions ($28 \times 28, 346 \times 260$), however, the effectiveness of our model on the event streams captured using high-resolution event cameras ($1280 \times 720$ or $1280 \times 800$) is still unclear. We leave the two points as our future works.

\section{Conclusion} \label{sec:conclusion}
In this paper, we propose a novel point-voxel absorbing graph representation learning framework for event stream based recognition. Specifically, we first transform the event stream into a sparse event cloud and voxel grids for a joint representation. The dual representations achieve a better trade-off between performance and efficiency. Then, we build dual graphs on the two inputs and also introduce absorbing nodes into the graph for global information aggregation. The absorbing graph convolution networks (AGCN) are designed for structured feature learning and global feature aggregation, simultaneously. The AGCN addresses the issues of fragmented node feature learning and global classification feature aggregation in the previous event-based classification models. Finally, we concatenate the outputs from the dual branches for classification. Extensive experiments on multiple event-based classification benchmark datasets fully validated the effectiveness of our framework.

\bibliography{report} 
\bibliographystyle{spiebib} 

\end{document}